\def\year{2020}\relax
\documentclass[letterpaper]{article} 
\usepackage{aaai20}  
\usepackage{times}  
\usepackage{helvet} 
\usepackage{courier}  
\usepackage[hyphens]{url}  
\usepackage{graphicx} 
\usepackage[utf8]{inputenc} 
\usepackage[T1]{fontenc}    
\usepackage{booktabs}       
\usepackage{amsfonts}       
\usepackage{nicefrac}       
\usepackage{microtype}      
\usepackage{multirow}
\usepackage{fontawesome}
\usepackage{amssymb}
\usepackage{amsmath}
\usepackage{algorithm}
\usepackage[linesnumbered,ruled,algo2e]{algorithm2e}

\title{Diversity Transfer Network for Few-Shot Learning}

\newcommand*\samethanks[1][\value{footnote}]{\footnotemark[#1]}

\author{
Mengting Chen\textsuperscript{\rm 1}\thanks{Equal contribution.}\thanks{Mengting Chen was an intern of Horizon Robotics when working on this paper.},
Yuxin Fang\textsuperscript{\rm 1}\samethanks[1], 
Xinggang Wang\textsuperscript{\rm 1}\thanks{Corresponding authors.}, 
Heng Luo\textsuperscript{\rm 2}, 
Yifeng Geng\textsuperscript{\rm 2}, \\ 
\Large \textbf{
Xinyu Zhang, 
Chang Huang\textsuperscript{\rm 2}, 
Wenyu Liu\textsuperscript{\rm 1}\samethanks[3], 
Bo Wang\textsuperscript{\rm 3, \rm 4}}\\
\textsuperscript{\rm 1} School of EIC, Huazhong University of Science and Technology\enskip
\textsuperscript{\rm 2} Horizon Robotics, Inc.\enskip \textsuperscript{\rm 3} Vector Institute\enskip \textsuperscript{\rm 4} PMCC, UHN \\
\{mengtingchen, yuxin\_fang, xgwang, liuwy\}@hust.edu.cn \enskip  \{heng.luo, yifeng.geng, chang.huang\}@horizon.ai\\
xyzhang.me@foxmail.com \enskip 
bowang@vectorinstitute.ai
}

\begin{document}
\maketitle
\begin{abstract}
  Few-shot learning is a challenging task that aims at training a classifier for unseen classes with only a few training examples. The main difficulty of few-shot learning lies in the lack of intra-class diversity within insufficient training samples. To alleviate this problem, we propose a novel generative framework, Diversity Transfer Network (DTN), that learns to transfer latent diversities from known categories and composite them with support features to generate diverse samples for novel categories in feature space. The learning problem of the sample generation (i.e., diversity transfer) is solved via minimizing an effective meta-classification loss in a single-stage network, instead of the generative loss in previous works. 
  Besides, an organized auxiliary task co-training over known categories is proposed to stabilize the meta-training process of DTN. We perform extensive experiments and ablation studies on three datasets, i.e., \emph{mini}ImageNet, CIFAR100 and CUB. The results show that DTN, with single-stage training and faster convergence speed, obtains the state-of-the-art results among the feature generation based few-shot learning methods. Code and supplementary material are available at: \texttt{https://github.com/Yuxin-CV/DTN}.
\end{abstract}

\section{Introduction}
\label{Introduction}

Deep neural networks (DNNs) have shown tremendous success in solving many challenging real-world problems when a large amount of training data is available \cite{krizhevsky2012imagenet,Simonyan2014VeryDC,He2015DeepRL}. Common practice suggests that models with more parameters have the greater capacity to fit data and more training data usually provide better generalization ability. However, DNNs struggle to generalize given only a few training data while humans excel at learning new concepts from just a few examples \cite{bloom2000children}. Few-shot learning has therefore been proposed to close the performance gap between machine learner and human learner. 
In the canonical setting of few-shot learning, there are a training set $\mathcal{D}_{train}$ (seen, known) and a testing set $\mathcal{D}_{test}$ (unseen, novel) with \textit{disjoint} categories. Models are trained on the training set while tested in an $N$-way $K$-shot scheme \cite{vinyals2016matching} where the models need to classify the queries into one of the $N$ categories correctly when only $K$ samples of each novel category are given. This unique setting of few-shot learning poses an unprecedented challenge in fully utilizing the prior information in the training set $\mathcal{D}_{train}$, which corresponds to the known information or historical information of the human learner. Common approaches to address this challenge either learn a good metric for novel tasks \cite{snell2017prototypical,vinyals2016matching,sung2017learning} or train a meta-learner for fast adaptation \cite{pmlr-v70-finn17a,pmlr-v70-munkhdalai17a,ravi2017}.

Recently, the generation based approach is becoming an effective solution for few-shot learning \cite{Hariharan_2017_ICCV,NIPS2018_7549,Wang_2018_CVPR,NIPS2018_7504}, since it directly alleviates the problem of lacking training samples. We propose a Diversity Transfer Network (DTN)  for sample generation. In DTN, the offset between a random sample pair from the known category is composited with a support sample in the novel category in the latent feature space. Then, the generated features, as well as the support features, are averaged as the proxy of the novel category. At last, query samples are evaluated by the proxy.
Only if the generated samples follow the distribution of the real samples to be diverse, can the meta-classifier (i.e., the proxy) be robust enough to classify queries correctly.

In addition to the new sample generation scheme, we utilize an effective meta-training curriculum called OAT (Organized Auxiliary task co-Training), inspired by the auxiliary task co-training in TADAM \cite{NIPS2018_7352} and curriculum learning \cite{bengio2009curriculum}. OAT organizes auxiliary tasks and meta-tasks reasonably and effectively reduces training complexity. Experiments show that by applying OAT, our DTN converges much faster compared with the na\"ive meta-training strategy (i.e., meta-training from scratch), the multi-stage training strategy used in $\Delta$-encoder \cite{NIPS2018_7549} and the auxiliary task co-training strategy used in TADAM.

The main components of DTN are integrated into a single network and can be optimized in an end-to-end fashion. Thus, DTN is very simple to implement and easy to train. Our experimental results show that this simple method outperforms many previous works on a variety of datasets.

\section{Related Work}
\label{related}

\subsection{Metric Learning Based Approaches}
\label{related-metric}

Metric learning is the most common and straightforward solution for few-shot learning. An embedding function can be learned by a myriad of instances of known categories. Then some simple metrics, such as Euclidean distance \cite{snell2017prototypical} and cosine distance \cite{vinyals2016matching,Qiao_2018_CVPR,Qi_2018_CVPR}, are used to build nearest neighbor classifiers for instances in unseen categories. 
Furthermore, to model the contextual information among support images and query images, bidirectional LSTM and attention mechanism are adopted in Matching Network \cite{vinyals2016matching}.
Besides measuring the distances of a query to its support images, there is a new solution that compares the query to the center of the support images of each class in feature space, such as \citeauthor{snell2017prototypical,Qiao_2018_CVPR,Qi_2018_CVPR,Gidaris_2018_CVPR}. The center is usually termed as a proxy of the class. Specifically, squared Euclidean distance is used in Prototypical Network \cite{snell2017prototypical}, and cosine distance is used in the other works. \citeauthor{snell2017prototypical,Qi_2018_CVPR} directly calculate proxies by averaging the embedding features, while \citeauthor{Qiao_2018_CVPR,Gidaris_2018_CVPR} take a small network to predict proxies.
Based on Prototypical Network, TADAM \cite{NIPS2018_7352} further proposes a dynamic task conditioned feature extractor by predicting the layer-level element-wise scale and shift vectors for each convolutional layer. Different from simple metrics, Relation Network \cite{sung2017learning} takes the neural network as a non-linear metric and directly predicts the similarities between the query and support images. 
TPN \cite{liu2018learning} performs transductive learning on the similarity graph contains both query and support images to obtain high-order similarities.




\subsection{Meta-Learning Based Approaches}
\label{related-meta}

Meta-learning approaches have been widely used in few-shot learning scenarios by optimization learning for fast adaptation, aiming to learn a meta-learner that can solve the novel task quickly. Meta Network \cite{pmlr-v70-munkhdalai17a} and adaResNet \cite{pmlr-v80-munkhdalai18a} are memory-based methods. Example and task level information in Meta Network are preserved in fast and slow weights, respectively. AdaResNet performs rapid adaptation by mimicking conditionally shifted neurons which modify activation values with task-specific shifts retrieved from a memory module. An LSTM-based update rule of the parameters of a classifier is proposed in \citeauthor{ravi2017}, where both short-term knowledge within a task and long-term knowledge common among all the tasks are learned. MAML \cite{pmlr-v70-finn17a}, LEO \cite{rusu2018meta} and MT-net \cite{pmlr-v80-lee18a} all differentiate through gradient update steps to optimize performance after fine-tuning. While MAML operates directly in high dimensional parameter space, LEO performs meta-learning within a low-dimensional latent space. Different from MAML that assumes a fixed model, MT-net chooses a subset of its weights to fine-tune. \citeauthor{pmlr-v80-franceschi18a} propose a method based on bi-level programming that unifies gradient-based hyper-parameter optimization and meta-learning.

\subsection{Generation Based Approaches}
\label{related-generation}

Sample synthesis using the generative models has recently emerged as a popular direction for few-shot learning \cite{Zhu_2017_ICCV,goodfellow2014generative}. How to synthesize new samples based on a few examples remains an interesting open problem. AGA \cite{Dixit_2017_CVPR} and FATTEN \cite{Liu_2018_CVPR} are attribute-guided (w.r.t. pose and depth) augmentation methods in feature space by leveraging a corpus with attribute annotations. \citeauthor{Hariharan_2017_ICCV} tries to transfer transformations from a pair of examples from a known category to a ``seed” example of a novel class. Finding specific generation targets requires a carefully designed pipeline with heuristic steps. $\Delta$-encoder \cite{NIPS2018_7549} also tries to extract intra-class deformations between image pairs sampled from the same class. \citeauthor{Wang_2018_CVPR} proposes to generate samples by adding random noises to support features. Different from previous methods, MetaGAN \cite{NIPS2018_7504} generates fake samples that need to be discriminated by the classifier instead of augmentation, which sharpens the decision boundaries of novel categories.  

Our proposed DTN shares a philosophical similarity with image hallucination \cite{Hariharan_2017_ICCV} and $\Delta$-encoder \cite{NIPS2018_7549} with distinct differences in the following aspects. The first difference is that DTN does not require to set specific target points for the generator. More specifically, $\Delta$-encoder takes a pair of images $X_a^1$ and $X_a^2$
from the same class and learns to infer the diversity between them by reconstructing $X_a^2$. The image hallucination method collects quadruples $(X_a^1, X_a^2, X_b^1, X_b^2)$ for training based on clustering and traversal; each quadruple contains two image pairs from two classes $a$ and $b$; a generation network is trained to predict a sample $X_a^2$ from the quadruple when the rest three $(X_a^1, X_b^1, X_b^2)$ are given as input. Then, synthesized samples are used to train a linear classifier. The input of the generator in DTN is also a triplet $(X_a^1, X_b^1, X_b^2)$ as \citeauthor{Hariharan_2017_ICCV}, but the generated sample $\hat{X}_a^2$ is used directly to construct the meta-classifier, and the generator is optimized by minimizing the meta-classification loss instead of setting specific generation targets. Secondly, DTN integrates feature extraction, feature generation, and meta-learning into a single network and enjoys the simplicity and effectiveness of end-to-end training, while \citeauthor{Hariharan_2017_ICCV,NIPS2018_7549} are stage-wise methods.

\begin{figure*}
  \centering
  \includegraphics[width=1.9\columnwidth]{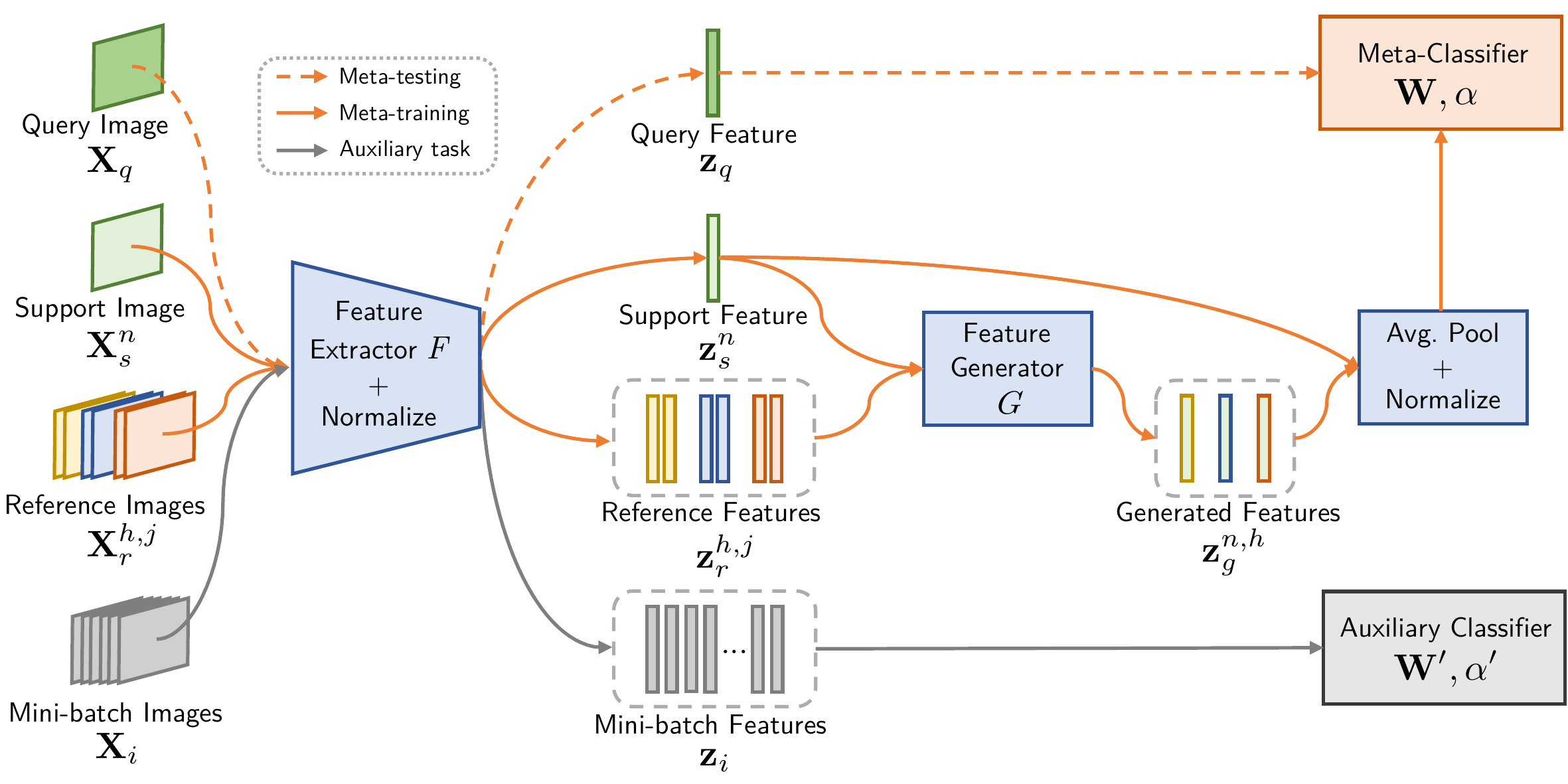}
\caption{\textbf{Illustration of the proposed diversity transfer network.} The branch indicated by orange arrows is the meta-task, which is trained in a meta-learning way. The orange solid arrows indicate the process of meta-training, while the orange dashed arrows indicate the process of meta-testing. During meta-training, the features of the support image and reference images from the feature extractor are fed into the feature generator to generate new features. The parameters of the meta-classifier are formed by the averaged proxies of the support features and generated features. Then the query image is fed to evaluate the performance of the meta-classifier during meta-testing. The branch indicated by grey arrows is the auxiliary task aimed to accelerate convergence and improve the generalization ability.}
  \label{fig-meta}
\end{figure*}

More recent work based on sample generation and data augmentation are IDeMe-Net \cite{chen2019deformation} and SalNet \cite{Zhang_2019_CVPR}. The former utilizes an additional deformation sub-network with a large number of parameters to synthesize diverse deformed images, the latter needs to pre-train a saliency network on the MSRA-B dataset. In contrast to these approaches, our method is based on a simple diversity transfer generator that learns a better proxy of each category with fewer parameters and faster convergence speed. Besides, our method can be regarded as an instance of compositional learning \cite{yuille2011towards} in the latent feature space.

\section{Method}
\label{method}

\subsection{Problem Definition}
\label{problem}

Different from the conventional classification task, where the training set $\mathcal{D}_{train}$ and the testing set $\mathcal{D}_{test}$ consist of samples from the same classes, few-shot learning aims to address the problem where the label spaces are disjoint between $\mathcal{D}_{train}$ and $\mathcal{D}_{test}$. We follow the standard $N$-way $K$-shot classification scenario defined in \citeauthor {vinyals2016matching} to study the few-shot learning problem. An $N$-way $K$-shot task is termed as an episode. An episode is formed by $N$ classes sampled from the training$/$testing set firstly. Then $K$ images sampled from each of the $N$ classes constitute the support set $\{(\mathbf{X}_s^{n,k}, Y_s^{n,k})\}$, where $n\in[1,2,...,N]$ and $k\in[1,2,...,K]$. For the sake of simplicity, we take $N$-way $1$-shot (i.e., $K=1$) classification for example in the following sections, and the support set will be simplified to $\{(\mathbf{X}_s^{n}, Y_s^{n})\}$. The query sample $(\mathbf{X}_q, Y_q)$ is sampled from the rest images of the $N$ classes. The goal is to classify the query into one of the $N$ classes correctly based only on the support set and the prior meta-knowledge learned from the training set $\mathcal{D}_{train}$.

\subsection{An Overview of Diversity Transfer Network}
\label{DTN}

The overall structure of the Diversity Transfer Network (DTN) is shown in Fig.~\ref{fig-meta}. DTN contains four modules and is organized into two task branches. The task branch indicated by orange arrows is the meta-task, which is trained in a meta-learning way. The input for the meta-task consists of the following three parts: support images $\mathbf{X}_s^{n}$, a query image $\mathbf{X}_q$ and reference images $\mathbf{X}_r^{h,j}$, where $n\in[1,2,...,N]$, $h\in[1,2,...,H]$, $j\in[1,2]$. All images are mapped to $L_2$-normalized feature vectors $\mathbf{z}=F(\mathbf{X})$ by a feature extractor $F$, where $\mathbf{z} \in \mathbb{R}^C$. $\mathbf{z}_r^{h,1}$ and $\mathbf{z}_r^{h,2}$ are feature vectors of two reference images. They come from the same category and make up a reference pair. The diversity of the pair is transferred to the support feature $\mathbf{z}_s^{n}$ to generated a new feature by the feature generator $G$. The generated feature $\mathbf{z}_g^{n,h}$ is supposed to belong to the same category with $\mathbf{z}_s^{n}$. For each support feature, there are $H$ samples generated based on it. Since a meta-task is an $N$-way $1$-shot image classification task, the meta-classifier is an $N$-way classifier consisting of a weight matrix $\mathbf{W}$ and a trainable temperature $\alpha$. The values in the $\mathbf{W}$ are determined by the proxies formed by support features and features generated by them. The meta-classifier is differentiable, so the feature extractor $F$ and feature generator $G$ can be updated by standard back-propagation according to the loss function defined by the cosine similarity between the query and the proxies. 
The task branch indicated by grey arrows in Fig.~\ref{fig-meta} is the auxiliary task, aiming to accelerate and stabilize the training of DTN. It is a conventional classification task over all categories of the training set $\mathcal{D}_{train}$. 

\begin{figure}
  \centering
  \includegraphics[width=1.0\columnwidth]{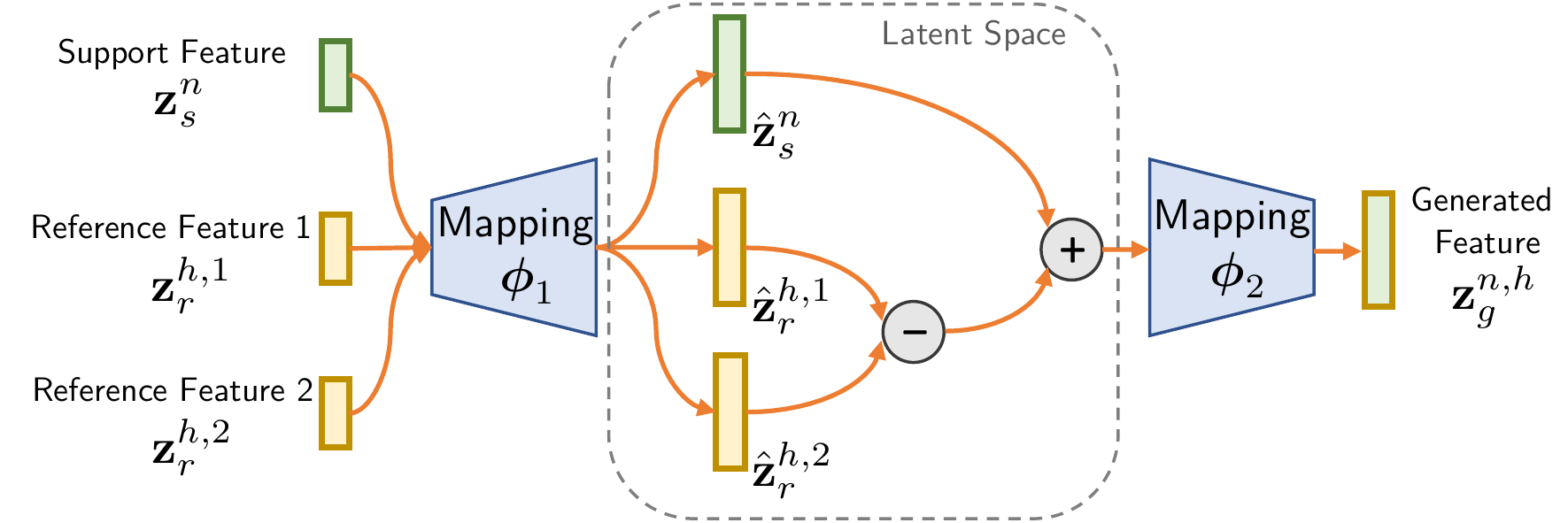}
\caption{\textbf{Feature generator in DTN.} The three input features are mapped into a latent space by the mapping function $\phi_1$. Then the diversity (i.e., offset) between the reference features is added with the support feature in this space. Then it is mapped by the $\phi_2$ to keep the same size as inputs. The output is a generated feature which is supposed to be a sample belonging to category $n$.}
  \label{fig:dtn}
\end{figure}

\subsection{Feature Generation via Diversity Transfer}
\label{generator}

Each image $\mathbf{X}$ is mapped to a feature vector $\mathbf{z}=F(\mathbf{X})$ by the feature extractor $F$. $\mathbf{z}_q$, $\mathbf{z}_s^n$ and $\{\mathbf{z}_r^{h, 1}, \mathbf{z}_r^{h, 2}\}$ are feature vectors of the query image $\mathbf{X}_q$, the support image $\mathbf{X}_s^n$ and the reference images pair $\{\mathbf{X}_r^{h,1}, \mathbf{X}_r^{h,2}\}$ respectively, where $n\in[1,2,...,N]$, $h\in[1,2,...,H]$. For a specific support feature $\mathbf{z}_s^n$, during both meta-training and meta-testing phase, the reference image pairs $\{\mathbf{X}_r^{h,1}, \mathbf{X}_r^{h,2}\}$ are always sampled from the training set $\mathcal{D}_{train}$ (seen, known).
 Specifically, we first randomly sample $H$ classes from the whole training classes $\mathcal{D}_{train}$ with replacement. For each sampled class, we then randomly sample two different images $\mathbf{X}_r^{h,1}$ and $\mathbf{X}_r^{h,2}$ to form a reference pair. We do not sample any images from $\mathcal{D}_{test}$ (unseen, novel) during the whole process.
The conventional few-shot evaluation setting, termed as $N$-way $K$-shot setting, requires to get a $N$-way classifier with the support of only $K$ samples for each novel class and the prior meta-knowledge from the whole training set $\mathcal{D}_{train}$. Therefore, our sampling method strictly complies with the few-shot learning protocol.
 
 As shown in Fig.~\ref{fig:dtn}, the feature generator $G$ of DTN consists of two mapping functions $\phi_1$ and $\phi_2$. Three input features are firstly mapped into a latent space $\hat{\mathbf{z}}=\phi_1(\mathbf{z})$, $\hat{\mathbf{z}}\in \mathbb{R}^{C'}$. The elementwise difference $\hat{\mathbf{z}}_r^{h,1}-\hat{\mathbf{z}}_r^{h,2}$ measures the diversity between the two reference features. It is applied to the support feature by a simple linear combination $\hat{\mathbf{z}}_s^n+(\hat{\mathbf{z}}_r^{h,1}-\hat{\mathbf{z}}_r^{h,2})$. After mapping it by $\phi_2$, we get a feature $\mathbf{z}_g^{n,h}$ which has the same size of the input $\mathbf{z}_s^{n}$ and should belong to the same category with the support feature $\mathbf{z}_s^n$. More specifically:


\begin{equation} \label{eq:1}
\begin{aligned}
\mathbf{z}_g^{n,h}=& G(\mathbf{z}_s^n,\mathbf{z}_r^{h,1},\mathbf{z}_r^{h,2}) \\
 =& \phi_2(\phi_1(\mathbf{z}_s^n)+\phi_1(\mathbf{z}_r^{h,1})-\phi_1(\mathbf{z}_r^{h,2})).
\end{aligned}
\end{equation}

Given $H$ different reference pairs for a single support feature $\mathbf{z}_s^n$, there will be $H$ generated features that enrich the diversity of category $n$. They are helpful to construct a more robust classifier for unseen categories. When $K>1$, each of the $K$ support samples is taken as a ``seed'' and $H$ samples are generated based on it. Therefore, there will be $K$ support samples and $K\times H$ generated samples for each novel category.

\begin{table*}[!h]\small
  \caption{\textbf{Few-shot images classification accuracies on \emph{mini}ImageNet.} `-': not reported.}
  \label{results_mini}
  \centering
  \resizebox{1.8\columnwidth}{!}{
  \begin{tabular}{lcc|cc}
    \toprule
    Methods  & Ref. & Backbone   & $5$-way $1$-shot     & $5$-way $5$-shot \\
    \midrule
    Matching Network (\citeauthor{vinyals2016matching}) & NeurIPS'16 & 64-64-64-64 & $43.56\% \pm 0.84\%$ & $55.31\% \pm 0.73\%$ \\
    Meta-Learn LSTM (\citeauthor{ravi2017})  & ICLR'17 & 64-64-64-64 & $43.44\% \pm 0.77\%$ & $60.60\% \pm 0.71\%$ \\
    MAML (\citeauthor{pmlr-v70-finn17a}) & ICML'17 & 32-32-32-32 & $48.70\% \pm 1.84\%$ & $63.11\% \pm 0.92\%$ \\
    Prototypical Network (\citeauthor{snell2017prototypical}) & NeurIPS'17 & 64-64-64-64 & $49.42\% \pm 0.78\%$ & $68.20\% \pm 0.66\%$ \\
    Relation Network (\citeauthor{sung2017learning}) & CVPR'18 & 64-96-128-256 & $50.44\% \pm 0.82\%$ & $65.32\% \pm 0.70\%$ \\
    MT-net (\citeauthor{pmlr-v80-lee18a}) & ICML'18 & 64-64-64-64 & $51.70\% \pm 1.84\%$ & - \\
    MetaGAN$^{\blacklozenge}$ (\citeauthor{NIPS2018_7504}) & NeurIPS'18 & 64-96-128-256 & $52.71\% \pm 0.64\%$ & $68.63\% \pm 0.67\%$ \\
    \citeauthor{Qiao_2018_CVPR} & CVPR'18  & 64-64-64-64 & $54.53\% \pm 0.40\%$ & $67.87\% \pm 0.20\%$ \\
    \citeauthor{Gidaris_2018_CVPR} & CVPR'18  & 64-64-64-64 & $56.20\% \pm 0.86\%$ & $73.00\% \pm 0.64\%$ \\
    \textbf{DTN (Ours)}$^{\blacklozenge}$ & & 64-64-64-64 & $\mathbf{57.89\% \pm 0.84\%}$ & $\mathbf{73.28\% \pm 0.65\%} $ \\
    
    \midrule
    
    \citeauthor{Gidaris_2018_CVPR} & CVPR'18  & ResNet-12 & $55.45\% \pm 0.89\%$ & $70.13\% \pm 0.68\%$ \\
    adaResnet (\citeauthor{pmlr-v80-munkhdalai18a}) & ICML'18 & ResNet-12 & $56.88\% \pm 0.62\%$ & $71.94\% \pm 0.57\%$ \\
    TADAM (\citeauthor{NIPS2018_7352}) & NeurIPS'18 & ResNet-12 & $58.50\% \pm 0.30\%$ & $76.70\% \pm 0.30\%$ \\
    \citeauthor{Qiao_2018_CVPR} & CVPR'18  & WRN-28-10 & $59.60\% \pm 0.41\%$ & $73.74\% \pm 0.19\%$ \\
    STANet (\citeauthor{Yan2019ADA}) & AAAI'19 & ResNet-12 & $58.35\% \pm 0.57\%$ & $71.07\% \pm 0.39\%$ \\
    TPN (\citeauthor{liu2018learning}) & ICLR'19  & ResNet-12 & $59.46\%$ & $75.65\% $ \\
    LEO (\citeauthor{rusu2018meta}) & ICLR'19 & WRN-28-10  & $61.76\% \pm 0.08\%$ & $77.59\% \pm 0.12\%$ \\

    
    $\Delta$-encoder (\citeauthor{NIPS2018_7549})$^{\blacklozenge}$ & NeurIPS'18 & VGG-16 & $59.9\%$ & $69.7\% $ \\
    
    IDeMe-Net (\citeauthor{chen2019deformation})$^{\blacklozenge}$ & CVPR'19 & ResNet-18$^{\spadesuit}$ & $59.14\% \pm 0.86\%$ & $74.63\% \pm 0.74\%$ \\
    
    SalNet Intra-class Hal. (\citeauthor{Zhang_2019_CVPR})$^{\blacklozenge}$ & CVPR'19 & ResNet-101$^{\clubsuit}$& $62.22\% \pm 0.87\%$ & $77.95\% \pm 0.65\%$ \\
    
    \textbf{Deep DTN (Ours)}$^{\blacklozenge}$ & & ResNet-12 & $\mathbf{63.45\% \pm 0.86\%}$ & $\mathbf{77.91\% \pm 0.62\%}$ \\
    \bottomrule
  \end{tabular}}
  \resizebox{0.8\textwidth}{!}{
  \begin{tabular}{lll}
    $^{\blacklozenge}$ Generation based approaches & $^{\spadesuit}$ Using a deformation sub-network&
    $^{\clubsuit}$ Using a saliency network pre-trained on MSRA-B \\
  \end{tabular}}
\end{table*}

\subsection{Meta-Learning Based on Averaged Proxies}
\label{metaclassifier}

The meta-task branch of DTN is shown in Fig.~\ref{fig-meta} indicated by orange arrows. The orange solid arrows and dashed arrows indicate the process of meta-training and meta-testing, respectively. Each image $\mathbf{X}$ is mapped to a feature vector $\mathbf{z}=F(\mathbf{X})$.
Similar to \citeauthor{Qiao_2018_CVPR,Gidaris_2018_CVPR,Qi_2018_CVPR}, all the features here are $L_2$-normalized vectors (i.e., $\|\mathbf{z}\|_2=1$). The support feature $\mathbf{z}_s^n$ and all the $H$ reference feature pairs $\{\mathbf{z}_r^{h,1}, \mathbf{z}_r^{h,2}\} (h\in[1,2,...,H])$  are fed into the generator $G$ to generate $H$ new features $\mathbf{z}_g^{n,h}$ ($H$ is set to $3$ in Fig.~\ref{fig-meta} for example). So we get $H+1$ features for the $n$-th category. The meta-task is an $N$-way classification task, therefore the meta-classifier is represented by a matrix $\mathbf{W}\in \mathbb{R}^{N\times C}$, in which each row $\mathbf{w}_n \in \mathbb{R}^{C}$ can be viewed as a proxy \cite{Movshovitz-Attias_2017_ICCV} of the $n$-th category. 
After obtaining all the $H+1$ features for category $n$, the $n$-th row $\mathbf{w}_n$ of $\mathbf{W}$, termed as averaged proxy, is the $L_2$-normalized average of those features:
\begin{equation} \label{eq:2-1}
\hat{\mathbf{w}}_n= \frac{1}{H+1}\left(\mathbf{z}_s^n+\sum _{h=1}^H{\mathbf{z}_g^{n,h}}\right), 
\end{equation}



\begin{equation} \label{eq:2-2}
\mathbf{w}_n = \frac{\hat{\mathbf{w}}_n}{ \| \hat{\mathbf{w}}_n \|}.
\end{equation}

All the averaged proxies are also $L_2$-normalized vectors, so that the meta-classifier essentially becomes a cosine-similarity based classification model.
After constructing the meta-classifier, the $L_2$-normalized query feature $\mathbf{z}_q$ is fed into it for evaluation. The prediction $\mathbf{p}=\mathbf{z}_q\mathbf{W}^\top$ is the combination of $n$ classification scores $p_n=\mathbf{z}_q\mathbf{w}_{n}^\top$ of each category. To further increase stability and robustness when dealing with a large number of categories, we adopt a learnable temperature $\alpha$ in our meta-task loss as \citeauthor{Qi_2018_CVPR}, where $\alpha$ is updated by back-propagation during training. The meta-task loss $\mathcal{L}$ can be defined as follow:

\begin{equation} \label{eq:4}
\mathcal{L}=-\log{\frac{\exp{({\alpha\mathbf{z}_q\mathbf{w}_{Y_q}}^\top)}}{\sum _{n=1}^N{\exp{({\alpha\mathbf{z}_q\mathbf{w}_{n}}^\top)}}}}.
\end{equation}

\subsection{Organized Auxiliary Task Co-training}\label{atoat}

In order to accelerate the convergence of training and get better generalization ability, the meta-learning network in DTN is jointly trained with an auxiliary task. The auxiliary task is a conventional classification for all $N'$ categories in $\mathcal{D}_{train}$. It shares the same feature extractor $F$ with the meta-task branch. 
Different from the meta-classifier $\mathbf{W} \in \mathbb{R}^{N \times C}$ , which consists of the averaged proxies, the auxiliary classifier $\mathbf{W'} \in \mathbb{R}^{N'\times C}$  after the feature extractor $F$ are randomly initialized and updated via back-propagation. The mini-batch $\{(\mathbf{X}_i, Y_i)\}$ is randomly sampled from the training set $\mathcal{D}_{train}$, where $i\in[1,2,...,B]$, and $B$ is the batch size. The auxiliary task loss $\mathcal{L}'$ has the same form as the meta-task loss $\mathcal{L}$:

\begin{equation} \label{eq:5}
\mathcal{L}'=-\log{\frac{\exp{({\alpha'\mathbf{z}_i\mathbf{w}'_{Y_i}}^\top)}}{\sum _{n=1}^{N'}{\exp{({\alpha'\mathbf{z}_i\mathbf{w}'_{n}}^\top)}}}},
\end{equation}
\noindent
where $\mathbf{z}_i$ is one of the training features in the mini-batch, $\mathbf{w}'_{n} \in \mathbb{R}^{C}$ is the $n$-th row of $\mathbf{W'}$, and $\alpha'$ is learnable.

In TADAM \cite{NIPS2018_7352}, the auxiliary task is sampled with a probability that is annealed exponentially. We observe some positive effects from this training strategy compared with na\"ive meta-training and multi-stage training in our DTN. 

However, the inadequacy of this approach is: the randomness in both the \textit{frequency} and the \textit{order} of the two tasks affects the final result to some extent, and the distribution of auxiliary tasks are \textit{unpredictable} rather than annealed exponentially, especially when the number of training epochs is not very large. 
Another problem brought by the randomness is that it is hard to determine the training schedule, e.g., the learning rate, the number of training epochs, etc., since the permutation of auxiliary tasks and meta-tasks varies according to the random seed.
We empirically find that the stochastic auxiliary task co-training strategy used in TADAM results in a large fluctuation in the meta-classification accuracy (over $ 4\% $, see Table~\ref{fluctuation} for details) when using different random seeds. This randomness makes the choice of hyperparameters as well as the training schedule more difficult.

Therefore, we propose the OAT (Organized Auxiliary task co-Training) strategy, which organizes auxiliary tasks and meta-tasks in a more orderly and more reasonable manner. 
More specifically, there are two kinds of training epochs:
the auxiliary training epoch $\mathcal{A}$
and the meta-training epoch $\mathcal{M}$
. We select $T$ training epochs to form one training unit $\mathcal{U}$, the $i$-th training unit $\mathcal{U}_i$ has $\gamma_i$ meta-training epochs, and $(T - \gamma_i)$ auxiliary training epochs. The array of $\gamma_i$ is denoted as $\boldsymbol{\gamma} = \{\gamma_i\}_{i = 1}^L$, where $L$ is the total number of training units. Then the total number of training epochs is $T \times L$, and the whole training sequence $\mathcal{S}$ can be expressed as follow:
\begin{equation} \label{eq:6}
\mathcal{S} = \sum_{i = 1}^{L} \Big( \left( T - \gamma_i \right) \mathcal{A} + \gamma_i \mathcal{M} \Big). 
\end{equation}

By changing $T$ and $\boldsymbol{\gamma} 
$, we can obtain different training sequences arranged in different frequency and order, which is proven to be more manageable and effective compared with the training strategy used in TADAM. 
Intuitively, we would like to gradually add harder few-shot classification tasks into a series of simpler auxiliary classification tasks.
Therefore the setting of $T$ and $\boldsymbol{\gamma}$ is quite simple and straightforward.
We choose $T = 5$ and $\boldsymbol{\gamma} = [0, 0, 1, 1, 2, 2]$ for training DTN, though a more careful scheduling may achieve better performance. Therefore the whole training sequence $\mathcal{S}$ is organized as follow:
\begin{equation} \label{eq:6}
\mathcal{S} = 5\mathcal{A} \times 2 + (4\mathcal{A} + 1\mathcal{M}) \times 2 + (3\mathcal{A} + 2\mathcal{M}) \times 2. 
\end{equation}
\noindent


\begin{table}\small
  \caption{\textbf{The $\mathbf{5}$-way $\mathbf{1}$-shot/$\mathbf{5}$-way $\mathbf{5}$-shot images classification accuracies on CIFAR100 and CUB.} `-': not reported.}
  \label{results_other}
  \centering
  \resizebox{1\columnwidth}!{ 
  \begin{tabular}{l|cc}
    \toprule
    Methods     & CIFAR100     & CUB \\
    \midrule
    Nearest neighbor  & $56.1\% / 68.3\%$ & $52.4\% / 66.0\%$ \\
    Meta-Learn LSTM (\citeauthor{ravi2017})    & - & $40.4\% / 49.7\%$ \\
    Matching Network (\citeauthor{vinyals2016matching})  & $50.5\% / 60.3\%$ & $49.3\% / 59.3\%$ \\
    MAML (\citeauthor{pmlr-v70-finn17a})   & $49.3\% / 58.3\%$ & $38.4\% / 59.1\%$ \\
    $\Delta$-encoder (\citeauthor{NIPS2018_7549})  & $66.7\% / 79.8\%$ & $69.8\% / 82.6\% $ \\
    \textbf{Deep DTN (Ours)}  & $\mathbf{71.5\% / 82.8\%}$ & $\mathbf{72.0\% / 85.1\%} $ \\
    \bottomrule
  \end{tabular}
  }
\end{table}

Initially, the auxiliary tasks could be considered as a simpler curriculum\cite{bengio2009curriculum}, later they bring regularization effects to meta-tasks.
Ablation studies show that compared with the training strategy used in TADAM, DTN trained by OAT obtains better and more robust results with a faster convergence speed.

\begin{figure*}
  \centering
  \includegraphics[width=0.85\textwidth]{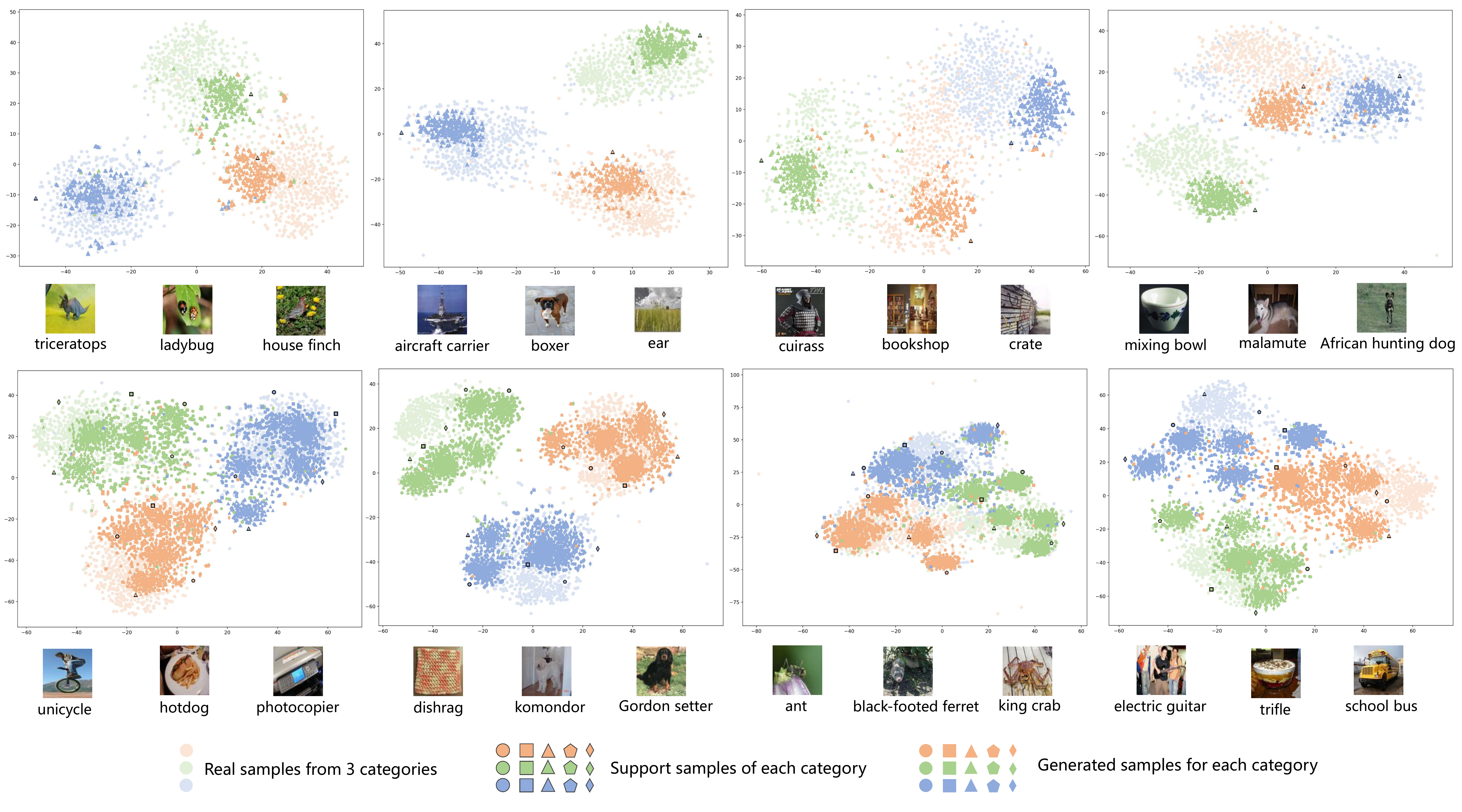}
\caption{\textbf{Visualization of generated samples, support samples, and real samples.} 
The light dots indicate real samples, the shapes (circle, square, triangle, pentagon and diamond) with black border indicate support samples which are also real samples, and the shapes without border indicate generated samples. There are $64$ generated samples for each support sample. The top row shows the results of $3$-way $1$-shot learning and the bottom row shows the results of $3$-way $5$-shot learning. The data in the left two columns are from the training set and the data in the right two columns are from the testing set.}
\label{fig-gen}
\end{figure*}

\section{Experiments}
\label{experiment}

\subsection{Implementation Details}
\label{implementation}

\noindent
\textbf{Dataset.} 
The proposed method is evaluated on multiple datasets: \emph{mini}ImageNet, CIFAR100 and CUB. The \emph{mini}ImageNet dataset has been widely used by few-shot learning since it is firstly proposed by \citeauthor{vinyals2016matching}. There are $64$, $16$ and $20$ classes for training, validation, and testing respectively. The hyper-parameters are optimized on the validation set. After that, it will be merged into the training set for the final results. The CIFAR100 dataset \cite{krizhevsky2009learning} contains 6000 images of 100 classes. We use 64, 16, and 20 classes for training, validation, and testing, respectively. The CUB dataset \cite{wah2011caltech} is a fine-grained dataset from 200 categories of birds. It is divided into training, validation, and testing sets with 100, 50, and 50 categories respectively. The splits of CIFAR100 and CUB follow \citeauthor{NIPS2018_7549}.

\noindent
\textbf{Architectures.} 
The feature extractor for DTN is a CNN with $4$ convolutional modules. Each module contains a $3\times 3$ convolutional layer with 64 channels followed by a batch normalization(BN) layer, a ReLU non-linearity layer, and a $2\times 2$ max-pooling layer. The structure of feature extractor is the same as those in former methods, e.g.,  \citeauthor{vinyals2016matching,snell2017prototypical} for fair comparisons. Many other works also use deeper networks for feature extraction to achieve better accuracy, e.g., \citeauthor{pmlr-v80-munkhdalai18a,NIPS2018_7352}. To make a comparison with them, we also implement our algorithm with ResNet-12 architecture\cite{He2015DeepRL}. The output of the feature extractor is a $1024$-dimensional vector. The mapping function $\phi_1$ in the feature generator $G$ is a fully-connected (FC) layer with $2048$ units followed by a leaky ReLU activation $(\max(x, 0.2x))$ layer, and a dropout layer with $30\%$ dropout rate. The mapping function $\phi_2$ has the same settings with $\phi_1$ except that the number of units of the FC layer is $1024$.

\begin{table*}[!h]\small
\centering 
  \caption{\textbf{Ablation studies on the fluctuation of the results obtained by AT and OAT on \emph{mini}ImageNet.} AT: auxiliary task co-training strategy used in TADAM.
OAT: organized auxiliary task co-training. 
The representation of the training sequence follows the notation introduced in the previous section, and the training sequence of AT is completely determined by the random seed. The results show that compared with the AT strategy(over $4\%$ fluctuation), the model trained by OAT (less than $1.5\%$ fluctuation) obtains better and more robust results.
}
  
  \label{fluctuation}
  \resizebox{1\linewidth}{!}{
  \begin{tabular}{c|c|cc|c|cc}
    \toprule
     \multirow{2}{*}{Random seed} & \multirow{2}{*}{\shortstack[c]{Training sequence of AT \\ (number of total training epochs = $30$)}} & \multicolumn{2}{c|}{Results of AT} & \multirow{2}{*}{\shortstack[c]{Training sequence of OAT\\ (number of total training epochs = $30$)}} & \multicolumn{2}{c}{Results of OAT} \\
      & & $5$-way $1$-shot & $5$-way $5$-shot & & $5$-way $1$-shot & $5$-way $5$-shot\\
    \midrule
    Seed $\#1$  & 13$\mathcal{A}$-1$\mathcal{M}$-2$\mathcal{A}$-1$\mathcal{M}$-1$\mathcal{A}$-1$\mathcal{M}$-2$\mathcal{A}$-1$\mathcal{M}$-2$\mathcal{A}$-1$\mathcal{M}$-5$\mathcal{A}$ & $58.60\%$ &  $72.78\%$ & 
    \multirow{5}{*}{10$\mathcal{A}$-4$\mathcal{A}$-1$\mathcal{M}$-4$\mathcal{A}$-1$\mathcal{M}$-3$\mathcal{A}$-2$\mathcal{M}$-3$\mathcal{A}$-2$\mathcal{M}$}
     &$62.78\%$ & $77.58\%$\\
    
    Seed $\#2$  & 11$\mathcal{A}$-1$\mathcal{M}$-7$\mathcal{A}$-1$\mathcal{M}$-7$\mathcal{A}$-3$\mathcal{M}$ & $60.71\%$ &  $74.10\%$ &  &$62.19\%$ & $76.82\%$\\
    
    Seed $\#3$  & 13$\mathcal{A}$-1$\mathcal{M}$-9$\mathcal{A}$-1$\mathcal{M}$-4$\mathcal{A}$-2$\mathcal{M}$ & $61.61\%$ &  $75.52\%$ &  &$63.45\%$ & $77.91\%$\\
    
    Seed $\#4$  & 18$\mathcal{A}$-1$\mathcal{M}$-2$\mathcal{A}$-1$\mathcal{M}$-1$\mathcal{A}$-1$\mathcal{M}$-2$\mathcal{A}$-2$\mathcal{M}$-1$\mathcal{A}$-1$\mathcal{M}$ & $60.97\%$ &  $75.77\%$ &  &$62.48\%$ & $77.22\%$\\
    
    Seed $\#5$  & 14$\mathcal{A}$-1$\mathcal{M}$-4$\mathcal{A}$-1$\mathcal{M}$-3$\mathcal{A}$-2$\mathcal{M}$-3$\mathcal{A}$-2$\mathcal{M}$ & $62.37\%$ &  $77.45\%$ &  &$63.17\%$ & $77.47\%$\\

    \bottomrule
  \end{tabular}
  }
\end{table*}

\begin{table}[!h]\small
  \caption{\textbf{Ablation studies for different feature generators on \emph{mini}ImageNet.}}
  \label{ablation_feat}
  \centering
  \resizebox{1\linewidth}{!}{
  \begin{tabular}{c|l|cc}
    \toprule
     & Methods    & $5$-way $1$-shot     & $5$-way $5$-shot \\
    \midrule
    
    
    
    1 & Gaussian noise generator
    & $60.14\% \pm 1.24\% $ & $75.57\% \pm 0.96\% $ \\
    
    2 & $\Delta$-encoder$^{\dagger}$
    & $60.38\% \pm 1.12\%$ & $73.44\% \pm 0.92\%$ \\
    
    3 & DTN w/ two-stage training & $61.95\% \pm 0.85\% $ & $76.52\% \pm 0.65\% $ \\
    
    4 & \textbf{DTN w/ OAT (Ours)} & $\mathbf{63.45\% \pm 0.86\%}$ & $\mathbf{77.91\% \pm 0.62\%} $ \\
    
    
    
    \bottomrule
  \end{tabular}}
  {
    $^{\dagger}$ Our reimplementation, which outperforms the original.
  }
\end{table}

\begin{table}[!h]\small
  \caption{\textbf{Ablation studies for different training strategies on \emph{mini}ImageNet.}  
  AT: auxiliary task co-training strategy used in TADAM 
  . OAT: organized auxiliary task co-training.}
  \label{ablation_tra}
  \centering
  \resizebox{1\linewidth}{!}{
  \begin{tabular}{c|l|cc}
    \toprule
     & Methods    & $5$-way $1$-shot     & $5$-way $5$-shot \\
    \midrule
    1 & DTN w/ na\"ive meta-training & $59.81\% \pm 1.36\% $ & $74.97\% \pm 1.01\% $ \\
    
    
    
    
    
    2 & DTN w/ two-stage training & $61.95\% \pm 0.85\% $ & $76.52\% \pm 0.65\% $ \\
    
    3 & DTN w/ AT
    & $62.65\% \pm 0.86\%$ & $77.12\% \pm 0.65\%$ \\
    
    4 & \textbf{DTN w/ OAT (Ours)} & $\mathbf{63.45\% \pm 0.86\%}$ & $\mathbf{77.91\% \pm 0.62\%} $ \\
    
    \bottomrule
  \end{tabular}}
\end{table}

\subsection{Results}
\label{result}
\noindent
\textbf{Quantitative Results.}
Table~\ref{results_mini} provides comparative results on the \emph{mini}ImageNet dataset. All these results are reported with $95\%$ confidence intervals following the setting in \citeauthor{vinyals2016matching}. Under the 4-CONV feature extractor setting, our approach significantly outperforms the previous state-of-the-art works, especially in the $5$-way $1$-shot task. As for the comparisons with models using deep feature extractor, deep DTN also surpasses other alternatives in the $5$-way $1$-shot scenario and achieves very competitive results under the $5$-way $5$-shot setting. The results confirm that our feature generation method is extremely useful to address the problem of learning with scarce data, i.e., the $5$-way $1$-shot scenario. DTN is also one of the simplest and lightweight feature generation methods which learns to enrich intra-class diversity, and does not rely on any extra information from other datasets, such as the salient object information in \citeauthor{Zhang_2019_CVPR}.


Table~\ref{results_other} shows that DTN also gets large improvements on the CIFAR100 and CUB datasets compared with existing state-of-the-arts in both $5$-way $1$-shot task and $5$-way $5$-shot task, which confirms DTN is generally useful for different few-shot learning scenarios.


\noindent
\textbf{Visualization Results.} 
To better understand the results, Fig.~\ref{fig-gen} shows tSNE \cite{maaten2008visualizing} visualizations of generated samples, support samples and real samples. It can be seen that our method can greatly enrich the diversity of an unseen class with only a single or a few support examples given. Most of the generated samples fit the distribution of real samples, which means that the category information of each support sample is well preserved by the generated sample, and they are close to the center of the real distribution even when the support sample lies on the edge. From the diagrams of $3$-way $5$-shot learning, it can be seen that generated features from $5$ support samples can cover the major distribution of the real samples, which facilitates to build a more robust classifier for unseen classes.

\subsection{Ablation Study}\label{ablation}
In this section, we study the impact of the feature generator, the training strategy and the number of generated features. We conduct the following ablation studies using models with deep feature extractor on \emph{mini}ImageNet. All the results are summarized in Table~\ref{fluctuation}, Table~\ref{ablation_feat}, Table~\ref{ablation_tra} and Table~\ref{dtn_diff_gen}.


First, we make a comparison between different feature generators. For the sake of fairness, we use exactly the same meta-classifier (cosine-similarity based classifiers) and the same training strategy (two-stage training), only the feature generators are different. All models are trained until convergence. Experiments show that diversity transfer generator outperforms Gaussian noise seeded generator (by $1.81\%$ in $5$-way $1$-shot, $0.95\%$ in $5$-way $5$-shot. Table~\ref{ablation_feat}, Row $1$ and Row $3$) and $\Delta$-encoder (by $1.57\%$ in $5$-way $1$-shot, $3.08\%$ in $5$-way $5$-shot. Table~\ref{ablation_feat}, Row $2$ and Row $3$).

Second, we study the effects of different training strategies. 
Obviously, OAT (Table~\ref{ablation_tra}, Row $4$) surpasses the na\"ive meta-training (Table~\ref{ablation_tra}, Row $1$) and two-stage training (Table~\ref{ablation_tra}, Row $2$). As mentioned before, a large fluctuation was observed (e.g., from $58.60\%$ to $62.37\%$ for $5$-way $1$-shot, from $72.78\%$ to $77.45\%$ for $5$-way $5$-shot, see Table~\ref{fluctuation} for details) in the meta-classification accuracy if the DTN is trained with auxiliary task sampled via a probability in 30 epochs. The result becomes better and more stable if we increase the total number of training epochs to 60 (Table~\ref{ablation_tra}, Row $3$), but this is still worse than the result obtained via DTN trained with OAT in only 30 epochs (Table~\ref{ablation_tra}, Row $4$). A comparison in the result's fluctuation between two training strategies is detailed in Table~\ref{fluctuation}. 

Finally, in Table~\ref{dtn_diff_gen} we study the impact on the number of generated features. The results gradually become better as the number of generated features increases. No improvement was observed when the number of generated features exceeds 64. We attribute this to the fact that 64 generated features have been well fitted to the real sample distribution.

\begin{table}\small
\centering
  \caption{\textbf{Ablation studies for DTN trained with different number of generated features on \emph{mini}ImageNet.} Numbers in the "( )" are difference in meta-classification accuracies compared with the result with $64$ generated features.}
  \label{dtn_diff_gen}
  \resizebox{0.9\columnwidth}!{ 
  \begin{tabular}{c|cc}
    \toprule
     Number of  & $5$-way & $5$-way \\
     generated features & $1$-shot & $5$-shot \\
    \midrule
    $2$    & $60.71\%$ $(-2.74\%)$ & $75.56\%$ $(-2.35\%)$\\
    $4$    & $61.18\%$ $(-2.27\%)$ & $76.17\%$ $(-1.74\%)$\\
    $16$   & $61.87\%$ $(-1.58\%)$ & $76.76\%$ $(-1.15\%)$\\
    $32$   & $62.58\%$ $(-0.87\%)$ & $77.14\%$ $(-0.77)\%$\\
    
    $\mathbf{64}$   & $\mathbf{63.45}\boldsymbol{\%}$ $(-0.00\%)$ & $\mathbf{77.91}\boldsymbol{\%}$ $(-0.00\%)$\\

    $96$   & $63.09\%$ $(-0.36\%)$ & $77.45\%$ $(-0.46\%)$\\
    $128$  & $63.26\%$ $(-0.19\%)$ & $77.99\%$ $(+0.08\%)$\\
    \bottomrule
  \end{tabular}}
\end{table}

\section{Conclusion and Future Work}
\label{conclusion}

In this work, we propose a novel generative model, Diversity Transfer Network (DTN), for few-shot image recognition. It learns transferable diversity from the known categories and augments the unseen category with the sample generation.
DTN achieves competitive performance on three benchmarks. We believe that the proposed generative method can be utilized in various problems challenged by the scarcity of supervision information, e.g., semi-supervised learning, active learning and imitation learning.
These interesting research directions will be explored in the future. 

\section{Acknowledgment}
\label{Acknowledgment}

This work was supported by National Natural Science Foundation of China (NSFC) (No. 61733007, No. 61572207 and No. 61876212), National Key R\&D Program of China (No. 2018YFB1402600) and  HUST-Horizon Computer Vision Research Center.

\small
\bibliographystyle{aaai}
\bibliography{AAAI-ChenM.4037}
\end{document}